\documentclass[3p,preprint]{elsarticle}

\usepackage{lineno}
\modulolinenumbers[5]

\newcommand{\eg}{{\em e.g.}}
\newcommand{\ie}{{\em i.e.}}

\makeatletter
\newcommand*{\rom}[1]{\expandafter\@slowromancap\romannumeral #1@}
\makeatother

\usepackage{graphicx}
\usepackage{amsfonts}
\usepackage{amssymb}
\usepackage{amsmath}

\usepackage[version=4]{mhchem}
\usepackage{siunitx}
\usepackage{longtable,tabularx}
\usepackage{bm}
\usepackage{multirow}
\usepackage{multicol}
\usepackage[colorinlistoftodos]{todonotes}
\usepackage[colorlinks=true, allcolors=blue]{hyperref}

\hypersetup{
    colorlinks,
    linkcolor={red!50!black},
    citecolor={blue!50!black},
    urlcolor={blue!80!black}
}

\journal{Computer-Aided Design}

\makeatletter
\def\@author#1{\g@addto@macro\elsauthors{\normalsize%
    \def\baselinestretch{1}%
    \upshape\authorsep#1\unskip\textsuperscript{%
      \ifx\@fnmark\@empty\else\unskip\sep\@fnmark\let\sep=,\fi
      \ifx\@corref\@empty\else\unskip\sep\@corref\let\sep=,\fi
      }%
    \def\authorsep{\unskip,\space}%
    \global\let\@fnmark\@empty
    \global\let\@corref\@empty  
    \global\let\sep\@empty}%
    \@eadauthor={#1}
}
\makeatother

\makeatletter
\def\ps@pprintTitle{%
   \let\@oddhead\@empty
   \let\@evenhead\@empty
   \let\@oddfoot\@empty
   \let\@evenfoot\@oddfoot
}
\makeatother

\begin{document}

\begin{frontmatter}

\title{B\'ezierGAN: Automatic Generation of Smooth Curves from Interpretable Low-Dimensional Parameters}

\author{Wei Chen\corref{mycorrespondingauthor}}
\cortext[mycorrespondingauthor]{Corresponding author}
\ead{wchen459@umd.edu}

\author{Mark Fuge}

\address{University of Maryland, College Park, Maryland, 20742}

\begin{abstract}
Many real-world objects are designed by smooth curves, especially in the domain of aerospace and ship, where aerodynamic shapes (\eg, airfoils) and hydrodynamic shapes (\eg, hulls) are designed. To facilitate the design process of those objects, we propose a deep learning based generative model that can synthesize smooth curves. The model maps a low-dimensional latent representation to a sequence of discrete points sampled from a rational B\'ezier curve. We demonstrate the performance of our method in completing both synthetic and real-world generative tasks. Results show that our method can generate diverse and realistic curves, while preserving consistent shape variation in the latent space, which is favorable for latent space design optimization or design space exploration.
\end{abstract}


\end{frontmatter}


\section{Introduction}

Smooth curves are widely used in the geometric design of products, ranging from daily supplies like bottles and drinking glasses to engineering structures such as aerodynamic or hydrodynamic shapes. 
However, the process of selecting the desired design is complicated, especially for engineering applications where strict requirements are imposed. For example, in aerodynamic or hydrodynamic shape optimization, generally three main components for finding the desired design are: (1)~a shape synthesis method (\eg, B-spline or NURBS parameterization), (2)~a simulator that computes the performance metric of any given shape, and (3)~an optimization algorithm (\eg, genetic algorithm) to select the design parameters that result in the best performance~\cite{kostas2017shape,yasong2018global}. We will review previous work on the first component\textemdash curve synthesis\textemdash in Section~\ref{sec:curve_synthesis}. A commonly used curve synthesis method in the design optimization domain is through different types of parameterization. However, there are two issues regarding this method: (1)~one has to guess the limits of the design parameters, thus the set of synthesized shapes usually cannot represent the entire pool of potential designs; (2)~the design space dimensionality is usually higher than the underlying dimensionality for representing sufficient shape variability~\cite{chen2017design}. 

While abundant design data (\eg, the UIUC airfoil database) has been accumulated today, useful knowledge can be inferred from those previous designs to facilitate the design process. Our proposed method learns a generative model from existing designs, and can generate realistic shapes with smooth curves from low-dimensional \textit{latent variables}. The generative model automatically infers the boundary of the design space and captures the variability of data using the latent representation. Thus it solves the above mentioned issues. Besides, to allow smooth exploration of the latent space, we regularize the latent representation such that shapes change consistently along any direction in the latent space. This method can also be treated as a parameterization method, where the parametric function (\ie, the generative model) is learned in a data-driven manner and usually more flexible to generate a wider range of potential designs, comparing to traditional parameterization methods like spline curves.



\section{Related Work}

Our proposed method synthesizes smooth curves by using a generative adversarial network (GAN)~\cite{goodfellow2014generative} based model. Thus in this section, we review previous work in curve synthesis and show the basics of the GAN and the InfoGAN~\cite{chen2016infogan}, a variant of standard GANs.

\subsection{Curve Synthesis}
\label{sec:curve_synthesis}

Curve synthesis is an important component in aerodynamic or hydrodynamic shape (\eg, airfoils, hydrofoils, and ship hulls) optimization, where curves representing those shapes are synthesized as design candidates. Splines (\eg, B-spline and B\'ezier curves)~\cite{je2001optimized,chi2016overview}, Free Form Deformation (FFD)~\cite{yasong2018global}, Class-Shape Transformations (CST)~\cite{berguin2014dimensionality,grey2018active}, and PARSEC parameterization~\cite{sobieczky1999parametric,derksen2010bezier,grey2018active} are used for synthesizing curves in the previous work. Then the control points or parameters for these parameterizations are modified (usually by random perturbation or Latin hypercube sampling~\cite{yasong2018global}) during optimization to synthesize design candidates. As mentioned previously, these methods suffer from the problems of unknown design parameter limits and the high-dimensionality of the design space.
Our proposed data-driven method eliminates these issues by using a low-dimensional latent representation to capture the shape variability and parameter limits of real-world designs.

There are also studies on the reverse design problem where functional curves like airfoils are synthesized from functional parameters (\eg, the pressure distribution and the lift/drag coefficient)~\cite{KHARAL2012330,di2014evolutionary}. These methods use neural networks to learn the complicated relationship between the curve geometry and its corresponding functional parameters. While the goal and method in our work are different, we share the idea of using neural networks to learn parametric curves.

Researchers in computer graphics have also studied curve synthesis for domains such as computer games and movies. Methods developed for this application are usually example-based, where curves are synthesized to resemble some input curve by applying hand-coded rules and minimizing some dissimilarity objective~\cite{hertzmann2002curve,merrell2010example,lang2015markov}. Different from these methods, our work serves a different purpose by targeting automatic curve synthesis without the need of providing examples.

\subsection{Generative Adversarial Networks}

\begin{figure}
\centering
\includegraphics[width=0.7\textwidth]{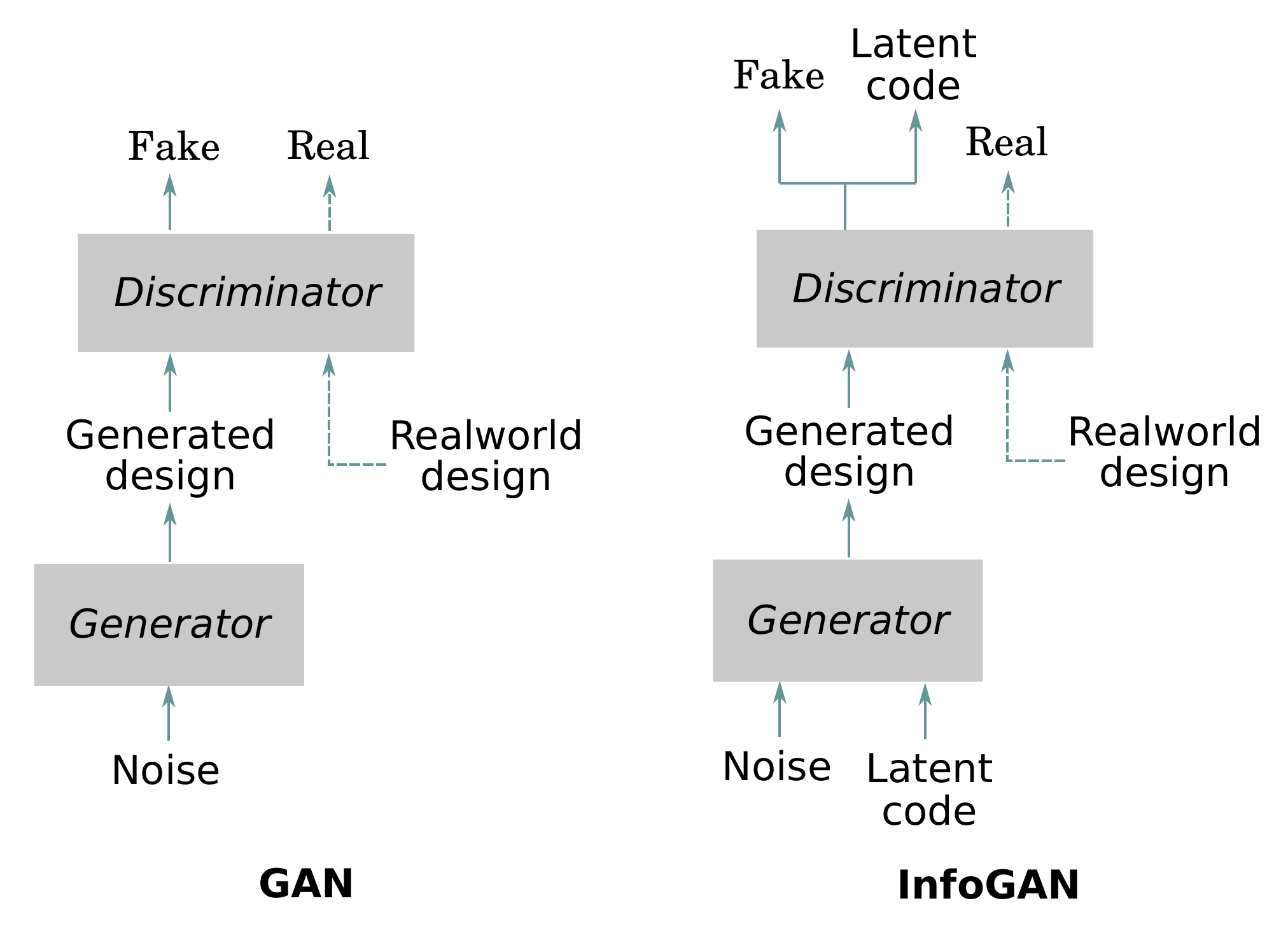}
\caption{Architectures of GAN and InfoGAN.}
\label{fig:gan}
\end{figure}

A generative adversarial network~\cite{goodfellow2014generative} consists of a generative model (generator $G$) and a discriminative model (discriminator $D$). The generator $G$ maps an arbitrary noise distribution to the data distribution (\ie, the distribution of curve designs in our scenario). The discriminator $D$ classifies between real-world data and generated ones (Fig.~\ref{fig:gan}). Both components improve during training by competing with each other: $D$ tries to increase its classification accuracy for distinguishing real-world data from generated ones, while $G$ tries to improve its ability to generate data that can fool $D$. The GAN's objective can be expressed as
\begin{equation}
\min_G\max_D V(D,G) = \mathbb{E}_{\bm{x}\sim P_{data}}[\log D(\bm{x})]+
\mathbb{E}_{\bm{z}\sim P_{\bm{z}}}[\log(1-D(G(\bm{z})))]
\label{eq:gan}
\end{equation}
where $\bm{x}$ is sampled from the data distribution $P_{data}$, $\bm{z}$ is sampled from the noise distribution $P_{\bm{z}}$, and $G(\bm{z})$ is the generator distribution. A trained generator thus can synthesize designs from a prior noise distribution. 

As for many curve synthesis problems, our goal is not just to generate curves, but also to facilitate design optimization and design space exploration by using a low-dimensional latent space to represent the geometrical design space. The noise input $\bm{z}$ can be regarded as a latent representation of the design space. However, $\bm{z}$ from the standard GAN is usually uninterpretable, meaning that the relation between $\bm{z}$ and the geometry of generated designs may be disordered and entangled. To mitigate this problem, the InfoGAN~\cite{chen2016infogan} uses a set of \textit{latent codes} $\bm{c}$ as an extra input to the generator, and regularizes $\bm{c}$ by maximizing a lower bound of the mutual information between $\bm{c}$ and the generated data. The mutual information lower bound $L_I$ is
\begin{equation}
L_I(G,Q) = \mathbb{E}_{\bm{x}\sim P_G}[\mathbb{E}_{\bm{c}'\sim P(\bm{c}|\bm{x})}[\log Q(\bm{c}'|\bm{x})]] + H(\bm{c})
\label{eq:li}
\end{equation}
where $H(\bm{c})$ is the entropy of the latent codes, and $Q$ is the auxiliary distribution for approximating $P(\bm{c}|\bm{x})$. We direct interested readers to~\cite{chen2016infogan} for the derivation of $L_I$. The InfoGAN objective combines $L_I$ with the standard GAN objective:
\begin{equation}
\min_{G,Q}\max_D V(D,G)-\lambda L_I(G,Q)
\label{eq:infogan}
\end{equation}
where $\lambda$ is a weight parameter. 

In practice, $H(\bm{c})$ is treated as constant if the distribution of $\bm{c}$ is fixed. The auxiliary distribution $Q$ is simply approximated by sharing all the convolutional layers with $D$ and adding an extra fully connected layer to $D$ to predict the conditional distribution $Q(\bm{c}|\bm{x})$. Thus as shown in Fig.~\ref{fig:gan}, the discriminator tries to predict both the source of the data and the latent codes $\bm{c}$~\footnote{Here we use the discriminator $D$ to denote both $Q$ and $D$, since they share neural network weights.}.

\section{B\'ezierGAN}

We propose a model, B\'ezierGAN, whose generator synthesizes sequences of discrete points on smooth curves. In this section, we introduce the architecture and optimization of this model.

\subsection{Overview}

\begin{figure}
\centering
\includegraphics[width=1.0\textwidth]{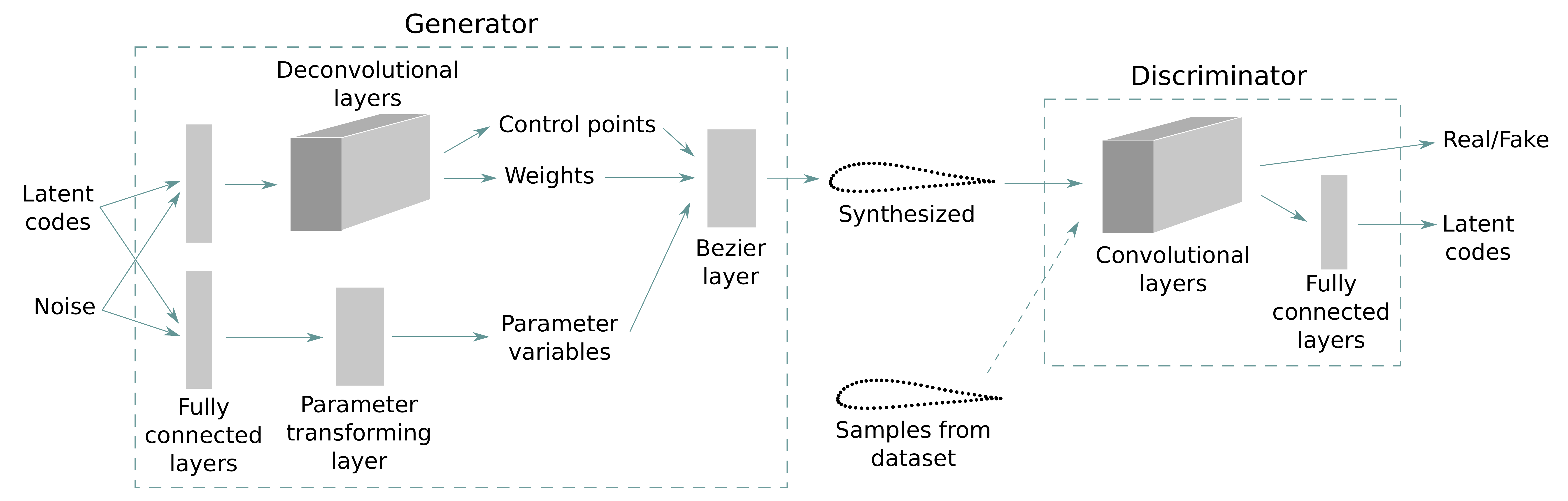}
\caption{Overall B\'ezierGAN model architecture.}
\label{fig:architecture}
\end{figure}

As shown in Fig.~\ref{fig:architecture}, the B\'ezierGAN adapts from the InfoGAN's structure. The discriminator takes in \textit{sequential discrete points} as data representation, where each sample is represented as a sequence of 2D Cartesian coordinates sampled along a curve. We omit the detailed introduction of the discriminator since it is the same as the one in the InfoGAN. For the generator, it converts latent codes and noise to control points, weights, and parameter variables of \textit{rational B\'ezier curves}~\cite{piegl1987interactive}, and then uses a B\'ezier layer to transform those B\'ezier parameters into sequential discrete points.

\subsection{Generating B\'ezier Parameters}

The latent codes and the noise are concatenated at the first layer, and go into two paths. On one path, control points $\bm{P}$ and weights $\bm{w}$ are generated through several fully connected and deconvolutioal layers. While on the other path, parameter variables $\bm{u}$ are generated through fully connected layers and a \textit{parameter transforming layer}. 

\begin{figure}
\centering
\includegraphics[width=0.5\textwidth]{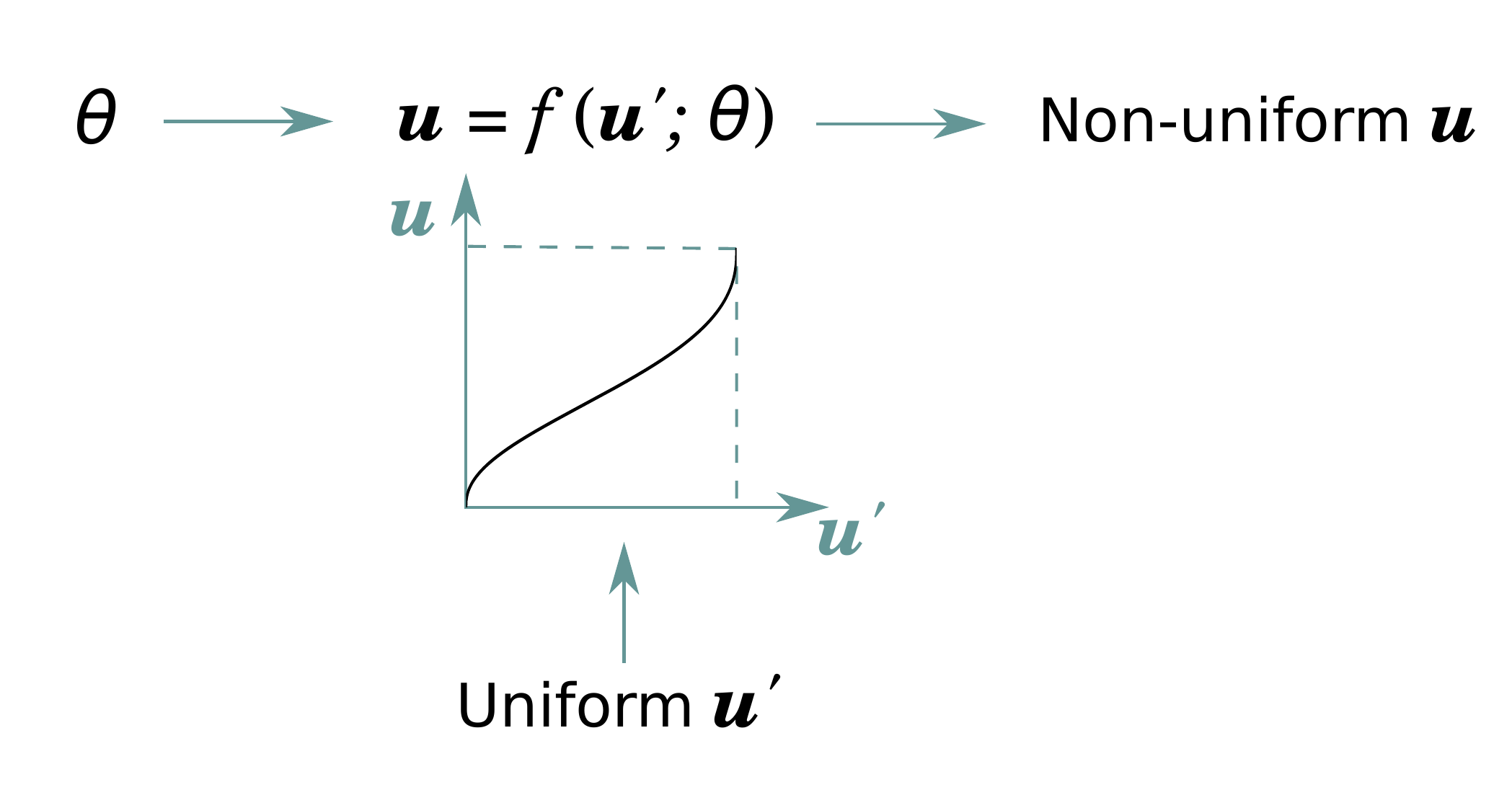}
\caption{Parameter transforming layer.}
\label{fig:parameter}
\end{figure}

The most straight forward way to sample along a B\'ezier curve is to use a uniform sequence of parameter variables $\bm{u}$. However, this makes our model less flexible and thus harder for the generator to converge. Also it is hard to directly learn $\bm{u}$ through fully connected or convolutional layers, since $\bm{u}$ has to be a sequence of increasing scalars. Thus we use a monotonically increasing function $f$ to convert a sequence of uniform parameter variables $\bm{u}'$ to non-uniform $\bm{u}$ (Fig.~\ref{fig:parameter}). Then $\bm{u}$ can be expressed as $\bm{u} = f(\bm{u}'; \theta)$, where the function parameters $\theta$ are obtained from the fully connected layers before the parameter transforming layer. 

We also want $\bm{u}'$ to be bounded and $\bm{u}$ us usually from 0 to 1. Thus a natural choice for $f$ will be the cumulative distribution function (CDF) of any distribution supported on a bounded interval. In this paper we use the CDF of Kumaraswamy’s distribution~\cite{jones2009kumaraswamy} due to its simple closed form. To make $f$ even more flexible, we set it to be the linear combination of a family of Kumaraswamy CDFs:
\begin{equation}
\bm{u} = \sum_{i=0}^M c_i(1-(1-(\bm{u}')^{a_i})^{b_i})
\label{eq:u}
\end{equation}
where $\bm{a}$ and $\bm{b}$ are parameters of the Kumaraswamy distributions, and $\bm{c}$ are the weights for Kumaraswamy CDFs and $\sum_{i=0}^M c_i = 1$ (which is achieved by a softmax activation). $M$ is the number of Kumaraswamy CDFs used. $\bm{a}$, $\bm{b}$, and $\bm{c}$ are learned from the fully connected layers.

\subsection{B\'ezier Layer}

The B\'ezier layer converts the learned B\'ezier parameters $\bm{P}$, $\bm{w}$, and $\bm{u}$ into sequential discrete points $\bm{X}$, based on the following expression~\cite{piegl1987interactive}:
\begin{equation}
\bm{X}_j = \frac{\sum_{i=0}^n\binom ni u_j^i(1-u_j)^{n-i}\bm{P}_i w_i}{\sum_{i=0}^n\binom ni u_j^i(1-u_j)^{n-i}w_i},~~~~j=0,...,m
\label{eq:bezier}
\end{equation}
where $n$ is the degree of the rational B\'ezier curve, and the number of discrete points to represent the curve is $m+1$. Since variables $\bm{P}$, $\bm{w}$, and $\bm{u}$ are differentiable in Eq.~\ref{eq:bezier}, we can train the network using regular back propagation.

\subsection{Regularization}

The B\'ezier representation (\ie, the choice of $\bm{P}$, $\bm{w}$, and $\bm{u}$) for a point sequence is not unique. For example, we have observed that the generated control points are dispersed and disorganized. The weights vanish at control points far away from the discrete points, and the parameter variables have to become highly non-uniform to adjust the ill-behaved control points. To prevent B\'ezierGAN from converging to bad optima, we regularize these B\'ezier parameters.

\paragraph{Control Points}  
Since the control points can be dispersed and disorganized, causing the weights and parameter variables to also behavior abnormally, one way to regularize control points is to keep them close together. We use the average and maximum Euclidean distance between each two adjacent control points as a regularization term:
\begin{align}
R_1(G) &= \frac{1}{Nn}\sum_{j=1}^N\sum_{i=1}^n \|\bm{P}_i^{(j)}-\bm{P}_{i-1}^{(j)}\| \\
R_2(G) &= \frac{1}{N}\sum_{j=1}^N\max_{i}\{\|\bm{P}_i^{(j)}-\bm{P}_{i-1}^{(j)}\|\}
\label{eq:reg_cp}
\end{align}
where $N$ is the sample size.

\paragraph{Weights}  
We use L1 regularization for the weights to eliminate the effects of unnecessary control points, so that the number of redundant control points is minimized:
\begin{equation}
R_3(G) = \frac{1}{Nn}\sum_{j=1}^N\sum_{i=0}^n |w_i^{(j)}|
\label{eq:reg_w}
\end{equation}

\paragraph{Parameter Variables}
To prevent highly non-uniform $\bm{u}$, we regularize parameters $\bm{a}$ and $\bm{b}$ from the Kumaraswamy distributions to make them close to 1, so that Eq.~\ref{eq:u} becomes $\bm{u} \simeq \bm{u}'$. The regularization term can be expressed as
\begin{equation}
R_4(G) = \frac{1}{NM}\sum_{j=1}^N\sum_{i=0}^M \|a_i^{(j)}-1\|+\|b_i^{(j)}-1\|
\label{eq:reg_ab}
\end{equation}

Then the objective of B\'ezierGAN is 
\begin{equation}
\min_{G,Q}\max_D V(D,G)-\lambda_0 L_I(G,Q)+\sum_{i=0}^4\lambda_i R_i(G)
\end{equation}

\subsection{Incorporating Symmetry}
\label{sec:symmetry}

There are cases where curves in the database are symmetric (axisymmetric, centrosymmetric, or rotational symmetric). A na\"ive way to generate them will be to just generate one part of the curve (we call it the \textit{prim}) using the B\'ezierGAN, and then obtain the rest by mirroring or rotating the prim as postprocessing. However, this na\"ive solution has a potential problem of neglecting the joints between the prim and the rest (this is shown in Fig~\ref{fig:sf}). Instead, the generator can first synthesize the B\'ezier parameters of the prim, and use symmetry or rotation operations to generate other B\'ezier parameters to obtain the full curve. Then the full curve can be feed into the discriminator, so that it will detect details at the joint and prevent the above problem. 

\paragraph{Axisymmetry} 
Given the prim's control points $\bm{P}$ and weights $\bm{w}$, we can obtain the control points $\bm{P}'$ and the weights $\bm{w}'$ of another axisymmetrical part by the following operations:
\begin{align}
\bm{P}' &= \bm{Q}\bm{P}\bm{S} \\
\bm{w}' &= \bm{Q}\bm{w}
\label{eq:symmetry}
\end{align}
where the permutation matrix $$\bm{Q}=
\begin{bmatrix}
    & & 1 \\
    & \reflectbox{$\ddots$} & \\
    1 & & 
\end{bmatrix},$$ and the symmetry matrix $$\bm{S}=
\begin{bmatrix}
    1 & 0 \\
    0 & -1
\end{bmatrix}$$ if the axis of symmetry is $x$, and
$$\bm{S}=
\begin{bmatrix}
    -1 & 0 \\
    0 & 1
\end{bmatrix}$$ if the axis of symmetry is $y$.

\paragraph{Rotational Symmetry or Centrosymmetry}
We deal with rotational symmetry and centrosymmetry using the same principle, since the latter is a special case of the former. To infer the B\'ezier parameters for other parts of the curve, we only need to rotate the prim's control points $\bm{P}$ and keep $\bm{w}$ fixed. The control points of a part rotational symmetrical to the prim is
\begin{equation}
\bm{P}' = \bm{P}\bm{R}
\end{equation}
where the rotation matrix $$\bm{R} = 
\begin{bmatrix}
    \cos\theta & -\sin\theta \\
    \sin\theta & \cos\theta
\end{bmatrix},$$
where $\theta$ is the rotation angle.

While we use the above operations for mirroring or rotating the control points or weights, the parameter variables are learned independently for each part of the curve, allowing for different sampling of points on each part.

\section{Experiments}

To evaluate our method, we perform generative tasks on four datasets. In this section, we describe our network and training configurations, and show both qualitative and quantitative results for the generated designs.

\subsection{Datasets}
We use two synthetic and two real-world design datasets. The synthetic datasets are created using superformula shapes. The real-world datasets are for aerodynamic design and hydrodynamic design, respectively.

\begin{figure}
\centering
\includegraphics[width=1\textwidth]{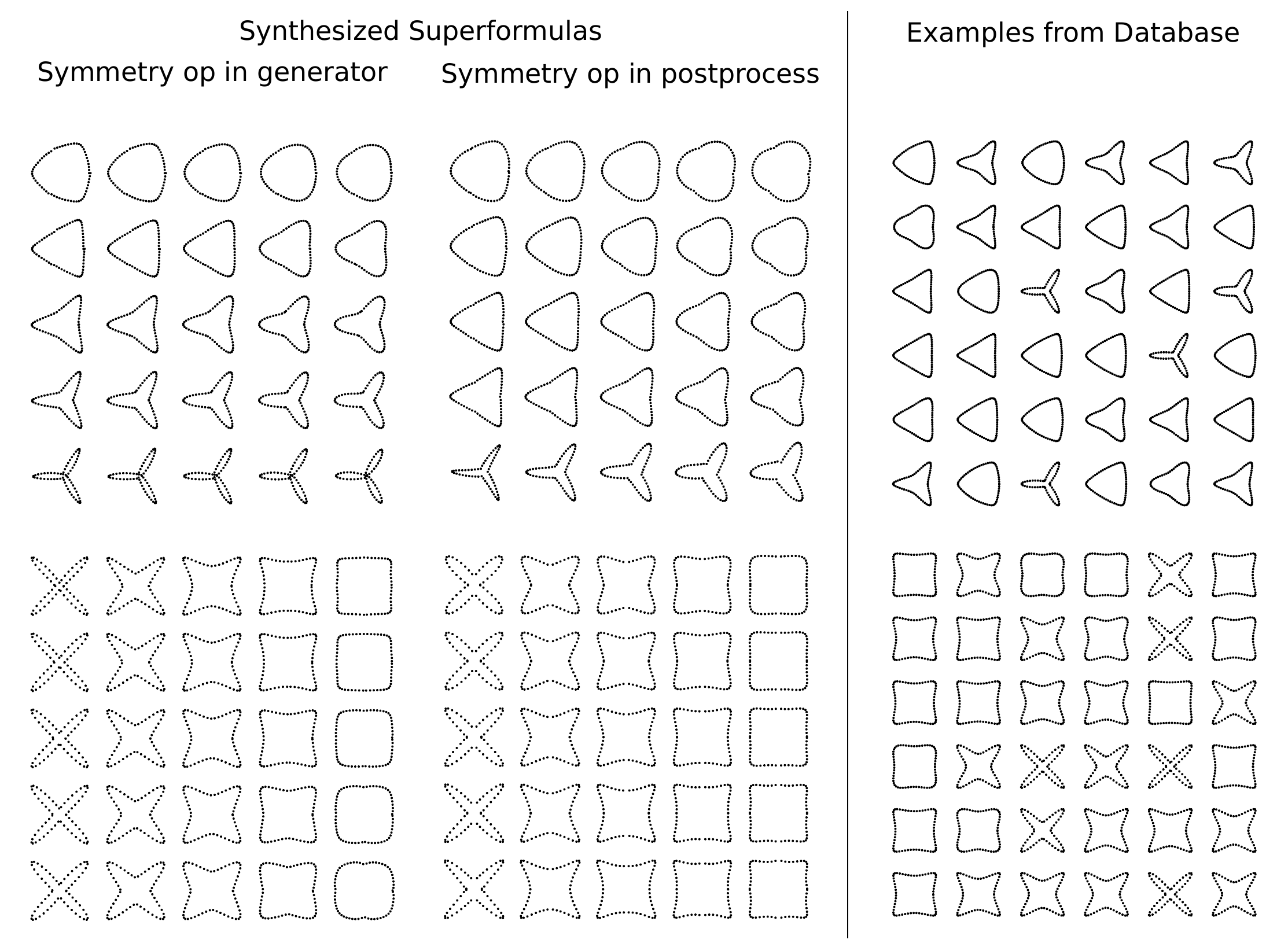}
\caption{Synthesized superformula shapes in 2-dimensional latent spaces. Examples in the created superformula dataset are shown on the right.}
\label{fig:sf}
\end{figure}

\begin{figure}
\centering
\includegraphics[width=1\textwidth]{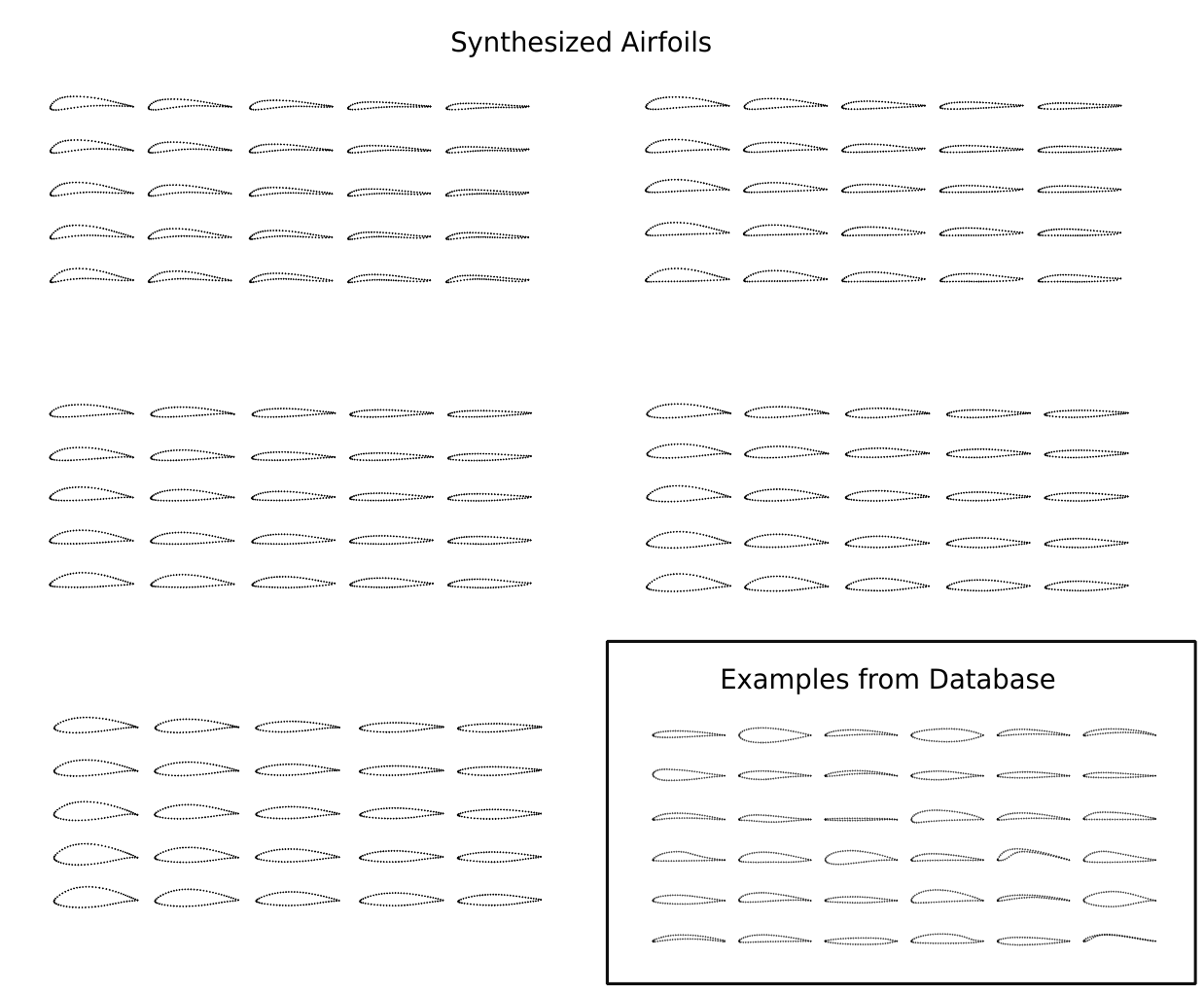}
\caption{Synthesized airfoil shapes in a 3-dimensional latent space. Examples from the airfoil database are shown at the bottom right.}
\label{fig:airfoil}
\end{figure}

\begin{figure}
\centering
\includegraphics[width=1\textwidth]{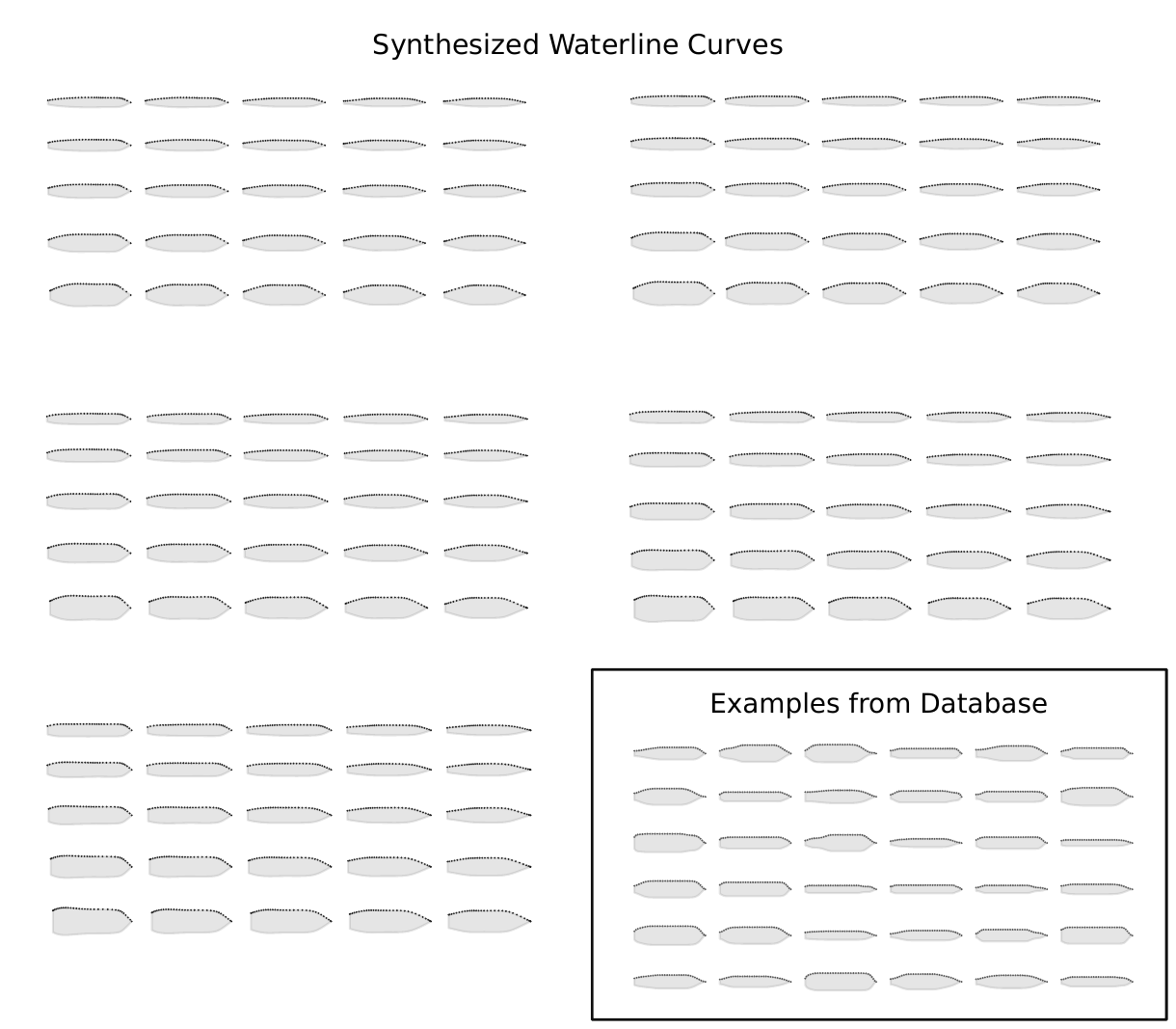}
\caption{Synthesized waterline curves in a 3-dimensional latent space. Examples from the created waterline curve dataset are shown at the bottom right.}
\label{fig:hull}
\end{figure}

\paragraph{Superformula}
As a generalization of the ellipse, superformula shapes are formed by periodic curves~\cite{gielis2003generic}. We generate two families of superformula shapes by using the following equations~\cite{chen2017design,chen2018hgan}:
\begin{equation}
\begin{split}
n_1 &= s_1 \\
n_2 &= n_3 = s_1+s_2 \\
r(\theta) &= \left(\left|\cos\left(\frac{m\theta}{4}\right)\right|^{n_2} + \left|\sin\left(\frac{m\theta}{4}\right)\right|^{n_3}\right)^{-\frac{1}{n_1}} \\
(x, y) & = (r(\theta) \cos\theta, r(\theta) \sin\theta)
\end{split}
\label{eq:sf}
\end{equation} 
where $s_1,s_2 \in [1,10]$, and $(x, y)$ is a Cartesian coordinate. For each superformula, we sample 64 evenly spaced $\theta$ from $0$ to $2\pi$, and get a sequence of 64 Cartesian coordinates. We set $m=3$ (Superformula~\rom{1}) and $m=4$ (Superformula~\rom{2}) respectively to get two families of superformula shapes (Fig.~\ref{fig:sf}). We control the shape variation of each superformula family with parameters $s_1$ and $s_2$.

\paragraph{Airfoil} 
We build the airfoil dataset by using the UIUC airfoil database\footnote{\url{http://m-selig.ae.illinois.edu/ads/coord_database.html}}. It provides the geometries of nearly 1,600 real-world airfoil designs, each of which is represented with sequential discrete points along its upper and lower surfaces.
The number of coordinates for each airfoil is inconsistent across the database, so we use B-spline interpolation to obtain a consistent shape representation. Specifically, we interpolate 64 points for each airfoil, and the concentration of these points along the B-spline curve is based on the curvature~\cite{je2001optimized} (Fig.~\ref{fig:airfoil}).

\paragraph{Hull Waterline Curve}
We use the Hull Lines Generator\footnote{\url{http://shiplab.hials.org/app/shiplines/}} to generate 2,000 design waterline (DWL) curves. A waterline curve represents the boundary between the underwater and the emerged portions of the hull, and is essential in ship design. The dataset only presents the upper half of each waterline since it is symmetric. We also interpolate 64 points using B-spline interpolation the same way as for the airfoil dataset (Fig.~\ref{fig:hull}).

\subsection{Model Configurations}

Each data sample is represented by a $64 \times 2$ matrix. In the discriminator, we use four layers of 1-dimensional convolution along the first axis of each sample. Batch normalization, leaky ReLU activation, and dropout are followed after each convolutional layer. We set the strides to be 2, and the kernel sizes 5. The depths of the four convolutional layers are \{64, 128, 256, 512\}. Fully connected layers are added after these layers to get the data source and latent codes prediction, as shown in Fig.~\ref{fig:architecture}.

For the generator, we first concatenate its two inputs: latent codes and noise. Then we use a fully connected layer and multiple 1-dimensional deconvolutional layers~\cite{zeiler2010deconvolutional} to output the control points and the weights. Separate fully connected layers are used to map inputs into parameters $\bm{a}$, $\bm{b}$, and $\bm{c}$. Specific configurations are different across datasets. Interested readers could check for detailed network architectures and hyperparameters in our Tensorflow implementation available on Github\footnote{\url{https://github.com/IDEALLab/bezier-gan}}. 

For the airfoil example, the last control point was set to be the same as the first control point, since the airfoil shape is a closed curve. For the waterline curve example, the last control point is at the head of the ship, and was set to be (1,0).

Since we only used two parameters to synthesize the superformula datasets, a latent dimension of two will be sufficient to capture the variability of superformula shapes. We set the latent dimension of the airfoil and the waterline curve examples as three. The latent codes are from uniform distributions, with each dimension bounded in the interval [0,1]. The input noise for each example is from a 10-dimensional multivariate Gaussian distribution.

\subsection{Training}

We optimize our model using an Adam optimizer~\cite{kingma2014adam} with the momentum terms $\beta_1=0.5$ and $\beta_2=0.999$. We set the learning rates of the discriminator and the generator to be 0.00005 and 0.0002, respectively. The batch size is 32, and the number of training steps is 5,000 for the airfoil and the waterline curve datasets, and 10,000 for the two superformula datasets.

The model was implemented using TensorFlow~\cite{tensorflow2015-whitepaper}, and trained on a Nvidia Titan X GPU. For each experiment, the wall-clock training time is shown in Table~\ref{tab:measure}, and the testing took less than 15 seconds.

\begin{table*}
\caption{Quantitative comparison between the B\'ezierGAN and the InfoGAN.}
\label{tab:measure}
\begin{center}
\begin{tabular}{ ccrrrr }
\hline
Example & Model & 
\multicolumn{1}{p{2cm}}{\centering MLL} & 
\multicolumn{1}{p{2cm}}{\centering RVOD} & 
\multicolumn{1}{p{2cm}}{\centering LSC} &
\multicolumn{1}{p{1.7cm}}{\centering Training \\ time (min)} \\
\hline
\multirow{2}{*}{Superformula \rom{1}} & B\'ezierGAN & $416.9 \pm 3.1$ & $0.933 \pm 0.001$ & $0.990 \pm 0.000$ & 13.26 \\
					 & InfoGAN & $410.0 \pm 2.0$ & $0.926 \pm 0.001$ & $0.983 \pm 0.000$ & 11.68 \\
\hline
\multirow{2}{*}{Superformula \rom{2}} & B\'ezierGAN & $432.9 \pm 3.1$ & $0.990 \pm 0.002$ & $0.968 \pm 0.001$ & 14.50 \\
					 & InfoGAN & $386.6 \pm 4.1$ & $0.961 \pm 0.002$ & $0.980 \pm 0.000$ & 11.77 \\
\hline
\multirow{2}{*}{Airfoil} & B\'ezierGAN & $260.3 \pm 5.8$ & $1.041 \pm 0.000$ & $0.952 \pm 0.002$ & 6.40 \\
					 & InfoGAN & $235.6 \pm 8.2$ & $0.674 \pm 0.000$ & $0.988 \pm 0.001$ & 6.29 \\
\hline
\multirow{2}{*}{Waterline Curve} & B\'ezierGAN & $198.0 \pm 1.5$ & $0.670 \pm 0.003$ & $0.981 \pm 0.001$ & 5.74 \\
					 & InfoGAN & $59.1 \pm 8.6$ & $0.203 \pm 0.000$ & $0.992 \pm 0.000$ & 5.84 \\
\hline
\end{tabular}
\end{center}
\end{table*}

\subsection{Visual Inspection}
\label{sec:visual}

\begin{figure}
\centering
\includegraphics[width=1\textwidth]{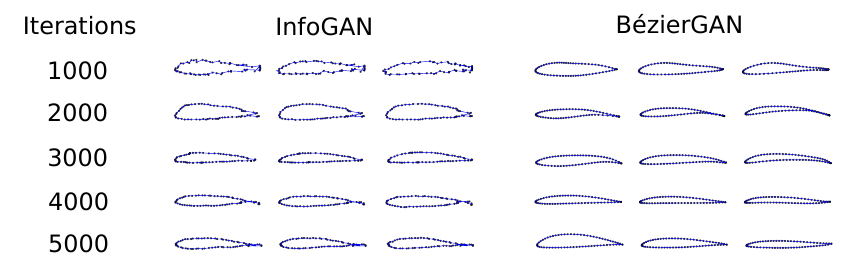}
\caption{Training processes for GANs with and without incorporating B\'ezier parameterization.}
\label{fig:iterations}
\end{figure}

To visualize and compare the training process between GANs with and without incorporating B\'ezier parameterization, we show generated shapes from every 1,000 training steps (Fig.~\ref{fig:iterations}). Results show that it is hard for the GAN without B\'ezier parameterization to generate smooth curves. Its generated curves are still noisy while B\'ezierGAN can generate realistic shapes.

The generated shapes in the latent space are visualized in Fig.~\ref{fig:sf}-\ref{fig:hull}. The plotted shapes are linearly interpolated in each latent space. Note that we visualize a 3-dimensional latent space by using multiple uniform slices of 2-dimensional spaces (Fig.~\ref{fig:airfoil} and \ref{fig:hull}). 

For both superformula examples, B\'ezierGAN captured pointiness and roundness of shapes, with each attribute varies along one dimension of the latent space (Fig.~\ref{fig:sf}). Incorporating symmetry conditions in the generator makes the shape as a whole look more realistic, and better captures the joints between parts than the na\"ive solution (\ie, apply symmetry conditions in postprocessing).

Figure~\ref{fig:airfoil} and \ref{fig:hull} show that the synthesized airfoils and waterline curves are realistic and capture most variation in their respective datasets. In the airfoil example, the horizontal axis captured upper surface protrusion, the vertical axis the roundness of the leading edge, and the third axis the lower surface protrusion of the overall airfoil shape. In the waterline curve example, the horizontal axis captured the length of the middle straight line, the vertical axis the width of the entire body, and the third axis the tail width.

\begin{figure}
\centering
\includegraphics[width=.7\textwidth]{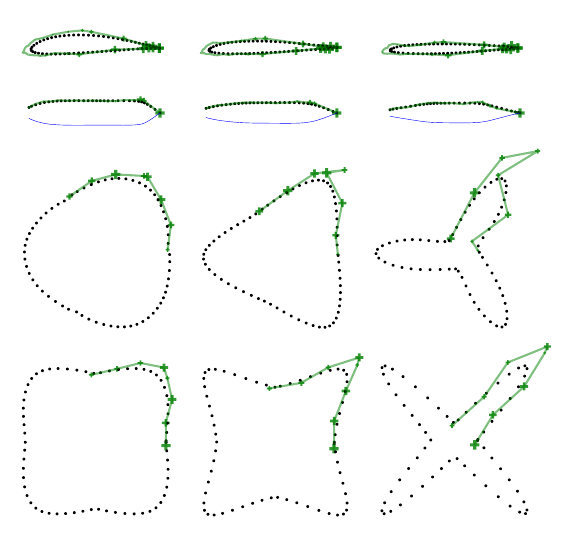}
\caption{Learned control points (``+'') and weights (indicated by the size of ``+'').}
\label{fig:control_points}
\end{figure}

The control points and weights of generated shapes are also visualized in Fig.~\ref{fig:control_points}. It shows reasonable control point positions, without deviating too much from the curves.

\subsection{Quantitative Evaluation}

Table~\ref{tab:measure} shows the quantitative performance measures and training time for each experiment.
The metrics are averaged over 10 test runs, each of which generates a different set of samples from the trained generator.

\paragraph{Test Likelihood}
The mean log likelihood (MLL) is a commonly used measure for generative models~\cite{breuleux2011quickly,goodfellow2014generative}. A high MLL indicates that the generative distribution well approximates the data distribution. It is determined by the likelihood of test data on the generative distribution. Table~\ref{tab:measure} shows that B\'ezierGAN had higher MLLs in all experiments.

\paragraph{Smoothness}
Though MLL is a good measure for generative quality, it does not explicitly capture smoothness, which is crucial in our curve synthesis task. Thus we use relative variance of difference (RVOD) to roughly measure the relative smoothness between our generated point sequences and those in the datasets. For a shape representing by a discrete points sequence $\bm{x}$, the variance of difference is expressed as
\begin{equation}
\text{VOD}(\bm{x}) = \frac{1}{m-1}\sum_{i=1}^{m-1}\text{Var}(\bm{x}_{i+1}-\bm{x}_i)
\end{equation}
where $m$ is the number of points in $\bm{x}$. Then RVOD can be expressed as
\begin{equation}
\text{RVOD} = \frac{\mathbb{E}_{\bm{x}\sim P_{data}}\text{VOD}(\bm{x})}{\mathbb{E}_{\bm{x}\sim P_G}\text{VOD}(\bm{x})}
\end{equation}
As expected, B\'ezierGAN outperforms InfoGAN regarding RVOD.

\paragraph{Latent Regularity}
Latent Space Consistency (LSC) measures the regularity of the latent space~\cite{chen2018hgan}. A high LSC indicates shapes change consistently along any direction in the latent space (\eg, a shape's roundness is monotonically increasing along one direction). This consistent shape change will results in a less complicated performance function and thus is beneficial for design optimization in the latent space. Since both B\'ezierGAN and InfoGAN use the mutual information loss to regularize the latent space, they both have high scores on LSC, with InfoGAN's LSC slightly higher in most experiments. This is expected since by adding additional regularization terms the model may trade off the original InfoGAN objective.

\section{Conclusion}

We introduced the B\'ezierGAN, a generative model for synthesizing smooth curves. Its generator first synthesizes parameters for rational B\'ezier curves, and then transform those parameters into discrete point representations. A discriminator will then exam those discrete points. The proposed model was tested on four design datasets. The results show that B\'ezierGAN successfully generates realistic smooth shapes, while capturing interpretable and consistent latent spaces.

Though this paper only demonstrated generative tasks for single curve objects, this method can generate more complex shapes with multiple curves by making the generator synthesize discrete points independently for each curve, and then concatenate synthesized points for all curves.

In spite of the complexity of those shapes, the real-world applications for 2D designs are limited. Thus we will also explore the possibility of generating smooth 3D surfaces using a similar model architecture.

\section*{Acknowledgment}
This work was supported by The Defense Advanced Research Projects Agency (DARPA-16-63-YFA-FP-059) via the Young Faculty Award (YFA) Program. The views, opinions, and/or findings contained in this article are those of the author and should not be interpreted as representing the official views or policies, either expressed or implied, of the Defense Advanced Research Projects Agency or the Department of Defense.


\bibliography{main}

\end{document}